# A Review of Vision-Based Assistive Systems for Visually Impaired People: Technologies, Applications, and Future Directions

Fulong Yao*, Wenju Zhou, Huosheng Hu

*Abstract*—Visually impaired individuals rely heavily on accurate and timely information about obstacles and their surrounding environments to achieve independent living. In recent years, significant progress has been made in the development of assistive technologies—particularly vision-based systems—that enhance mobility and facilitate interaction with the external world in both indoor and outdoor settings. This paper presents a comprehensive review of recent advances in assistive systems designed for the visually impaired, with a focus on state-of-the-art technologies in obstacle detection, navigation, and user interaction. In addition, emerging trends and future directions in visual guidance systems are discussed.

## I. Introduction

ACCORDING to the official report from World Health Organization (WHO), there are around 1.3 billion visually impaired people in the world [1]: approximately 36 million are completely blind and others have varying degrees of visual impairment. The First World Report on Vision [2], issued by the WHO, also released that the combination of growing and aging population will significantly increase the total number of people with eye conditions and vision impairment. The visual impairment has a dramatic impact on the individual's life, including his/her study, work, and daily communication. It is tough for visually impaired people to achieve the interaction with the surroundings without any help [3], especially in an unfamiliar environment, which is a breeze for ordinary people.

Traditional assistive tools for visually impaired people, such as the white cane and the guide dog, are no longer suitable for currently complex and changing situations [4]. Although the white cane can sense surrounding obstacles, it cannot detect the further object and suspended objects such as branches or bars. The guide dog is also used to help visually impaired people travel outside, even in unexplored environments. However, this method has some limits including high training costs and the relatively short service time. Also, it is quite difficult for the visually impaired people to take care of the dogs. As a matter of fact, almost all the traditional assistive tools cannot provide enough reference information for the blind, such as the shape of objects or the speed of moving objects, which limits users' interaction with the environment [5]. Moreover, about one in ten blind people use wheelchairs according to the literature [6]. For blind wheelchair users, it is difficult to independently travel by using a guide dog or the white cane. An ideal assistive system should not only provide blind people with information about obstacles, but also need to correlate their current location with features presented in the surrounding environment [7].

Recently, with the rapid development of sensing technology, more and more sensing-based assistive systems have been developed to help visual impairments interact better with others and complex environments. These systems have demonstrated greater adaptability and practicality than traditional assistive tools [8]. This paper reviews a lot of literature works in assistive systems for visually impaired people and attempts to present a comprehensive survey of the state-of-the-art technologies in this field. Different from the previous overviews of blind assistive systems [6], [9], this paper focuses on summarizing the latest assistive systems, especially visually assistive systems, from the hardware devices to the technologies (method and algorithm), as shown in Fig. 1. Various devices and technologies are compared in the paper, and their strengths and weaknesses are analysed in terms of cost, accuracy, and performance.

The rest of this paper is organized as follows. Section 2 overviews the assistive systems, and focuses on the explanation of traditional and electronically assistive systems. Section 3 introduces the hardware part of the visually assistive systems. Section 4 further describes the technology part (including advanced methods and algorithms) of the visually assistive systems. Some typical indoor and outdoor applications are presented in Section 5. Section 6 discusses the remaining challenges, while a brief conclusion is given in Section 7.

## II. Overview of Assistive Systems

Generally, the assistive system (AS) refers to some devices or items that is able to perceive the surroundings in real-time and helps individuals with visual impairment live like normal people. It has been a necessity for visually impaired people and will continue to be so in the coming years. Broadly, the AS can be classified into three

Corresponding author: Fulong Yao (yaofulong12345@gmail.com)

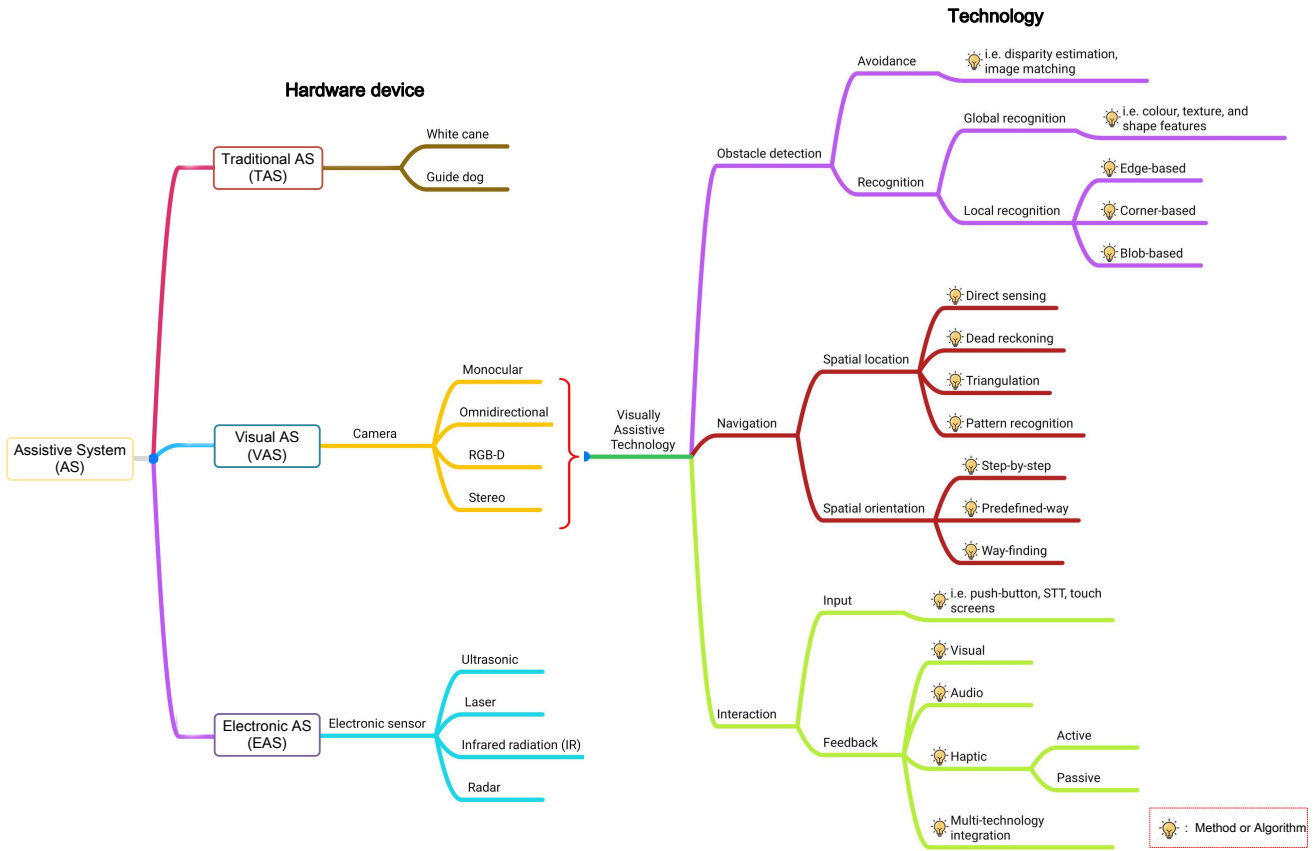

Fig. 1. The main framework of this paper.

categories according to the use of different technologies: traditional assistive system (TAS), electronically assistive systems (EAS), and visually assistive system (VAS). The TAS and EAS are reviewed in this section, and the VAS will be highlighted in the Section IV.

*A. Traditional Assistive System (TAS)*

The TAS can be roughly divided into two categories: white cane and guide dog. White cane, the most common TAS in the world, originated in the early 20th century and was known around the world in the mid-20th century [10]. It can immediately detect and avoid surrounding obstacles, as shown in Fig. 2 (a). But its detectable scope is limited by the cane length (usually only around 1.5 meters), and the users can only detect close objects on ground level. In other words, this tool can barely sense objects in suspension, since it senses the environment by touching and scanning the ground.

The guide dog, as shown in Fig. 2 (b), is another mature TAS for the blind, it can help the visually impaired to travel, even at traffic-intensive intersections. As a living individual, the guide dog not only can help the visually impaired in life, but also psychologically comfort them [11]. However, this method also has some constrains such as tough for dogs to understand the multiplex directions, and dogs can only service for a limited time (around 5 years) [12]. In addition, most blind individuals cannot afford a guide dog because of the long-term and complex training process with the high costs [13]. To make matters worse, guide dogs easily collide with other dogs in public, which may cause a terrible impact on the blind and even hurt them [14].

Both above TASs cannot adequately get the additional information such as shape or size of obstacles. These TASs only have the ability to detect some static obstacles, many other dynamic objects (cars, cats, etc.) are out of the reach. Furthermore, it is inconvenient for visually impaired people to hold the long cane in crowded places or take care of the dog in daily life.

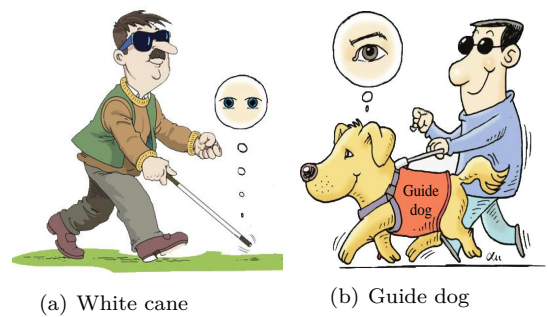

(a) White cane  (b) Guide dog

Fig. 2. Traditional assistive systems (TASs).

### B. Electronically Assistive System (EAS)

Visually impaired individuals always want to get more than just information about nearby obstacles. They prefer to relate the information to the features in the surroundings. Recently, the evolution in technology has made the trend for the EAS to replace the TAS. EAS, as the name implies, refers to the use of electronic sensors to capture information from the surroundings and then deliver the information to the visually impaired. It is designed to help visually impaired people perceive the complicated surroundings, and even achieve autonomous navigation in unfamiliar environments. The most representatively electronic sensors include ultrasonic, laser and infrared radiation (IR).

The detection principle of these sensors is like that of radar, shown in Fig. 3. The position and the direction of the obstacles can be obtained by calculating the transfer time of the photon/wave delivered from the transmitter to the receiver. Ultrasonic is a kind of electronic sensor which is often used in the EAS. Jaffer et al. developed a prebuilt ultrasonic based map to help visually impaired people travel around [15], while a novel electronic shoe called NavGuide was designed in [16] to provide the visually impaired people with obstacle free path-finding. However, this technology can't obtain the accurate direction, and is disturbed easily by other signals. The challenges on adopting ultrasonic measurements in uncertainty environments were further discussed in [17].

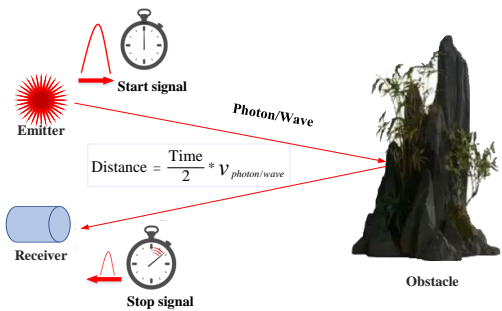

Fig. 3. Principle of obstacles detection for several common electronic sensors.

The laser-based EAS has been extensively studied in the past years to alleviate the negative effects of visual impairment. Dang et al. used the laser sensor as input to propose a virtual blind cane system for the blind to sense the obstacle and the distance [18]. However, the laser sensor is too expensive and cannot get the shape and volume of obstacles, which make laser rarely used in the wearable EAS. To solve this problem, Al-Fahoum et al. designed a smart infrared-based assistive system to help the visually impaired avoid nearby obstacles [19]. This AS is able to identify two significant features of obstacles, namely shape and the material. However, it cannot be applied to detect dynamic obstacles such as moving animals, and cannot determine the travel direction. Although these above EASs have greatly improved the travel capabilities of the visually impaired people, they cannot help the blind to find the optimal path or achieve purposeful navigation.

There are some electronic technologies, such as radio frequency identification (RFID), are often used to help the blind find out optimal paths or short-range automatic navigation [20]. Willis and Helal introduced an RFID-based navigation system for visually impaired people [21]. In this system, some RFID tags that can store about 2000 bits of data, including location and surrounding information, are placed in specific places. Users are able to freely navigate in indoors by recognizing RFID tags with the RFID reader attached on the cane or the shoe. Especially, the time for recognizing tags is short enough and not affecting normal walking. However, when the system is running, the distance between the RFID tags and the reader must be less than 100mm [22]. Moreover, the users have to detect two or more tags before they determine the specific forward direction, which limits the practicality of RFID. In addition, to reduce the user's load, Avila et al. designed a smart navigation system based on the small lightweight drone [23], which has a volume of 4.2 * 4.2 * 2 cm$^3$, and a weight of 29 g.

The invention of EAS has greatly improved the mobility of visually impaired people, since it can obtain a lot of scene information that TAS can't perceive. But compared to the wearable EAS, many blind people prefer to use a traditional white cane or a guide dog because it makes them feel safer [19]. Thus, some researchers combined the advantages of the EAS and TAS to develop some hybrid assistive system. Bolgiano et al. combined the laser with a white stick to enable visually impaired people to travel freely [24]. Poggi et al. integrated the visual sensor into a white cane and proposed an obstacle avoidance system based on the machine learning [25], whereas a guide dog robot was developed in [26] to enhance the mobility of the blind.

## III. Overview of Visual Sensors

VAS is one of ASs that adopts visual sensors to obtain the surrounding information. Vision, as the most used sense, can convey more information than olfactory, tactile, or auditory cues [27], [28]. According to the principles and structures, the types of vision sensors are different, mainly including monocular camera [29], omnidirectional camera [30], RGB-D camera [31], and stereo camera [32], as shown in Fig. 4.

### A. Monocular Camera

At the early stage of development, the VAS was normally designed with the monocular camera that capture the two-dimensional image, because it can not only detect suspended objects but also recognize dynamic obstacles[33]. In [34], a head mounted wide-angle monocular was adopted to help the blind to reach and grasp objects. Tapu et al. used the monocular camera embedded in the smartphone to design a real-time obstacle recognition system for the visually impaired people [35]. And a novel blind navigation system with a monocular camera was introduced in [36] to assist the blind in unknown dynamic scenes. However, the monocular camera cannot provide the accurate distance for users. Moreover, the limited visual range also restricts the development of this sensor.

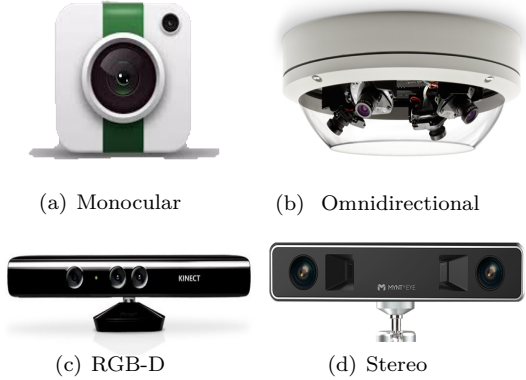

(a) Monocular  (b) Omnidirectional
(c) RGB-D  (d) Stereo

Fig. 4. Representative vision sensors.

### B. Omnidirectional Camera

Omnidirectional camera, also known as 360-degree camera, is another 2D camera widely used in the VAS. It usually contains multiple fixed cameras or a rotatable camera. Omnidirectional camera has a field of view that covers approximately the entire sphere or at least a full circle in the horizontal plane [37]. In [38], a first-person localization and navigation system with an omnidirectional camera was developed to help visually impaired travel freely. Cinaroglu et al. proposed an object detection method based on an omnidirectional camera [39], while a wearable omnidirectional vision system was introduced to assist visually impaired for autonomous navigation and recognition [40]. Although the omnidirectional camera can obtain a wider field of view than a monocular camera, it still does not get rid of the limitation from 2D images. As a matter of fact, object recognition by capturing 2D projections of the 3D world is inaccurate and even misleading, especially when it comes to multiple commodities [41].

### C. RGB-D Camera

With the development of sensing technology, the use of 3D vision to help blind people travel outside has become a new trend. Compared with 2D cameras, 3D cameras break the limitation of the distance by acquiring the depth information of the subject. The RGB-D camera is currently one of the most representative 3D sensors. It is a specific type of camera that is used in conjunction with the RGB camera, as shown in Fig.4 (c). And it can enhance traditional images with depth information (related to the distance to the sensor) on the pixel basis.

In [42], a wearable navigation system with an RGB-D camera is designed for visually impaired to extend the range of their activities. Neto et al. introduced a Kinect-based wearable face recognition system to help visually impaired people expand their social circle [43].

According to different measurement principles, RGB-D cameras can be split into two categories: structured light and time-of-flight (TOF), both of which are able to obtain directly depth information [41]. For a structured light camera such as Kinect V1, the camera projects a specific pattern onto the object, and calculates the depth or surface information of the object by sensing the deformation of the pattern through the infrared camera, as shown in Fig. 5. Although the RGB-D camera based on structured light can measure the distance information, it has the limitations of low accuracy, low spatial resolution, and easy motion blur. In the case of a TOF camera such as Kinect V2. The camera gets depth information by calculating the propagation time of the light pulse between the camera and the obstacle, and its principle is similar to the ultrasonic sensor, as shown in Fig. 3. Compared to the former sensor, the type of RGB-D camera has the advantage of high efficiency and accuracy. However, it is sensitive to the sunlight, which greatly limits its application in outdoors.

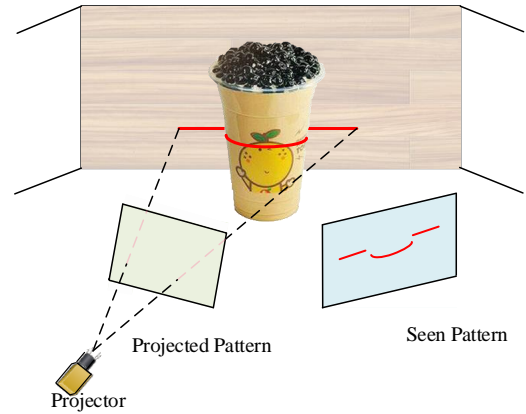

Fig. 5. Principle of RGB-D cameras based on structured light technology.

### D. Stereo Camera

The stereo camera, another type of 3D camera, has two or more lenses with a separate image sensor or film frame for each lens, as shown in Fig. 4(d). It captures three-dimensional images through simulating the human binocular vision. Generally, the camera has two matching lenses, and two pictures grabbed by two lenses can be projected simultaneously to create a three-dimensional impression, as shown in Fig. 6. Due to the different positions of the two cameras, a point in the real world is projected by the two cameras onto the two film frames. The points in the *Image*1 is shifted by a certain length in the *Image*2. When the relative position of each point in each camera is known (sometimes a tough task), the depth value $D$ can be calculated based on the pair of points with the following formula [41]:

$$D = F \times \frac{B}{d} \quad (1)$$

where $B$, as shown in Fig. 6, is the baseline length between the cameras, $F$ is the focal length of the two cameras, and $d$ is the disparity of the same point in different film frames. For Point 1, $d = d1 + d2$.

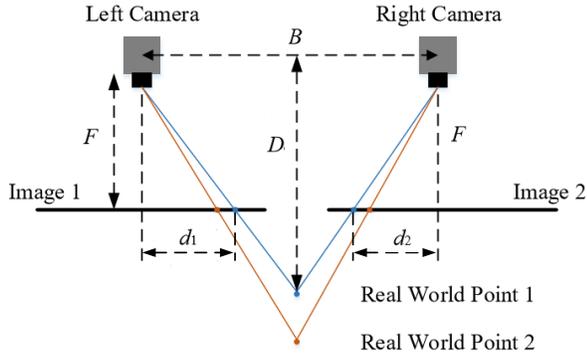

Fig. 6. Principle of stereo cameras.

The stereo camera has high accuracy and can work well in outdoor environment. For example, Kim et al. adopted a stereo camera as the visual inputs to design a novel virtual cane system that could be operated by an easy finger-pointing gesture [44]. Also, a navigation system based on the stereo camera is introduced in [45] to help the blind travel around in indoors or outdoors. However, these systems are sensitive to light and cannot be used at nearly distance case. Moreover, it has high computational complexity and poor real-time performance.

*E. Comparison of Visual Sensors*

Various visual sensors can be used to assist the visually impaired people to perceive and interact with surroundings. However, every devices comes with its own limitations, benefits, and costs. TABLE I shows the comparison of these visual sensors [46]. We can choose the proper sensor as the visual source for the specific purpose. Besides, there is no single, universal sensor or technology that aids in providing the necessary information in all situations. So a satisfactory assistive system may sometimes combine or integrate several visual sensors and even non-visual sensors.

## IV. VISUALLY ASSISTIVE SYSTEM (VAS)

The VAS for visually impaired individuals is consisted of two parts, the one is the hardware device, and the other is the advanced technologies. In the above sections, some common assistive devices have been introduced. In this section, some of the technologies or methods involved in VAS will be further discussed. A qualified assistive system for the visually impaired people must include these functions: obstacle avoidance and recognition, spatial location and orientation, and friendly interaction. Fig. 7 shows the function structure of the VAS.

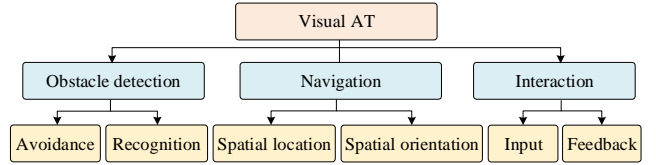

Fig. 7. Structure of the visually assistive system (VAS).

*A. Obstacle Detection*

There is a vast amount of literature on obstacle detection for blind people. According to different design purposes, the obstacle detection can be divided into two categories: Avoidance and Recognition.

*1) Obstacle Avoidance:* Obstacle avoidance is mainly dedicated to helping visually impaired avoid static or dynamic obstacles, and is widely used in blind outdoors or indoors travel. It usually uses the method of measuring the distance to detect obstacles and alert users. In the early days, the most common way was to use specific sensors, such as radar [15], [17], [47], laser [18], IR [19], to directly measure the distance. The principle of this method is shown in Fig. 3. Later, more and more people began to use visual cues to obtain distance information as the rise of visual sensing technology [48], especially the 3D camera. For example, Sekhar et al. adopted a real-time disparity estimation algorithm to evaluate the distance of obstacles [49]. Anderson et al. introduced an ID signal matching algorithm for stereo vision correlation to get the distance of obstacles [50]. However, both algorithms are not suitable for low-texture or low-light situations. Moreover, they fail to provide visually impaired with more information about obstacles such as shape, volume, materials, etc.

*2) Obstacle Recognition:* In daily life, it is not enough for the visually impaired to just avoid the surrounding obstacles, they need more information about obstacles. For example, when the blind person wants to find a chair, the single distance information is obviously not enough, and they also need information about the shape, size, and direction of the chair. Obstacle recognition is designed for such cases. It is dedicated to helping the blind to interact better with the environment and to live like normal people, even in strange environments. In [19], Al-Fahoum et al. developed an infrared-based intelligent blind guidance system that can obtain the shapes and materials of regular obstacles by measuring the width of the echo pulse signal. However, this system is not suitable for obstacles with complex shapes or materials, and it cannot detect the colour and texture characteristics of obstacles. Fortunately, visual sensors can almost describe all the elementary characteristics by the relationship among pixels. In [25], a robust RANSAC framework is proposed to discriminate between obstacles and the safe plane on which the user can walk onto. But the system only focused on the recognition of the obstacle that the user will meet first. According to different feature extraction, vision-based obstacle recognition can be classified into two categories: global recognition and local recognition [51].

TABLE I
Comparison of visual sensors

|  | Monocular | Omnidirectional | Stereo | RGB-D(TOF) | RGB-D(Structured light) |
| --- | --- | --- | --- | --- | --- |
| Software complexity | Low | Low | High | Low | High |
| Material cost | Low | Low | Low | Middle | High/Middle |
| Response time | Fast | Fast | Middle | Fast | Slow |
| Low light | Weak | Weak | Weak | Good | Depends on light source |
| Outdoor | Good | Good | Good | Fair | Weak |
| Depth accuracy | - | - | cm | mm to cm | um to cm |
| Depth Range | - | - | Middle range | Short range (<1m) to long range ( 40m) | Short range (cm) to middle range (4-6m) |

Global recognition usually uses global features extracted from the statistics of all the pixels, i.e. colour, texture, and shape features, to describe an image as a whole without considering the spatial layout in the image [41]. This type of recognition is mostly dedicated to large-scale detection, such as lawns, roads, and buildings. For example, Angin et al. designed a traffic light detector for blind navigation [52]. The system recognized the real-time status of traffic lights in video frames by adopting a cascade of boosted classifiers based on the AdaBoost algorithm and haar-like features [53]. The approach has the characteristics of good invariance and intuitive representation, but their high feature dimensions and a large amount of calculation are their serious weaknesses. Moreover, it is not suitable for cluttered background and occlusion. Therefore, its performance is degraded when it is applied to an indoor environment where the image background is complicated.

Local recognition often uses local features, which represent images by image patches, to present obstacles. According to the principles of local features, local recognition approach can be further classified into three categories: edge-based approach, corner-based approach, and blob-based approach [54]. Fig. 8 shows the principles of local features in computer vision [55].

The edge is a set of consecutive pixels whose brightness changes significantly in the digital image. Edge-based approach generally extracts the object contour from complex backgrounds by detecting the edges of objects. It can greatly reduce the amount of data, and remove information that can be considered irrelevant, retain important structural properties of the image. Currently, edge detectors, such as Canny [55], Sobel, Prewitt, and Roberts [56], are only used to generate some complementary information to construct and recognize objects' features. For example, Fernandes et al. proposed a blind navigation system based on stereo vision, in which the Canny edge detector was used to extract the region of interest to pave the way for subsequent feature extraction [45].

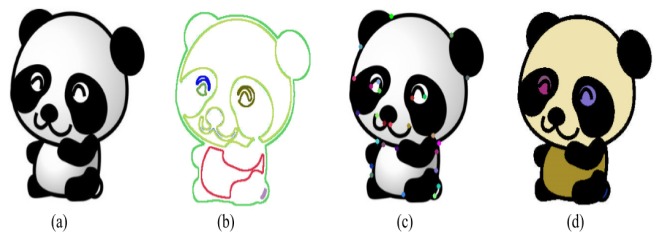

Fig. 8. Principles of local features in computer vision :(a) Original image;(b) Edge;(c) Corner; (d) Blob [41].

A corner, also named corner point, is usually defined as the intersection of two edges. More strictly speaking, the local neighbourhood of a corner point should have two different areas with different direction boundaries. Corner-based approach refers to detecting corner points with specific features to extract objects whose points have specific coordinates in the image and have certain mathematical properties such as local maximum value and gradient features. This approach can cope with changes in lighting and rotation to some extent. Currently, corner-based approaches, such as Harris [57] and FAST [58], are widely used in feature extraction and object matching. Kanayama et al. proposed a camera-computer system to help blind people walk safely [59]. In this system, the Harris corner detector is employed to detect pedestrians or moving objects.

A blob is a region of connected pixels with same or similar characteristics such as intensity and colour. Blob-based approach is dedicated to identifying and segmenting the target region from complex backgrounds by comparing its characteristics to their neighbouring regions. Blob-based approaches such as MSER (maximally-stable extremal region) [60] are mainly used in object recognition and region extraction. Deshpande and Shriram developed a real-time text detection and recognition system for blind people, in which the text is detected using MSER feature [61]. Although MSER is highly computational efficiency and robust to affine transformations, it is only suitable for distinguishing objects with small piexls changes from high-contrast scenes.

There are also some systems that use a combination of local and global features to recognize unknown objects and achieve surprising recognition results. In [62], a clothes recognition system based on vision was designed to recognize clothes colours and patterns for blind people. In this paper, global features of clothing patterns are obtained by the wavelet subbands, while scale invariance feature transform (SIFT) is adopted to extract local features.

To help the visual impairments, someone works on the vision-based marker recognition that identifies objects by scanning known markers such as quick response (QR) codes and barcodes. The marker-based recognition can recognize markers and detect obstacles with low power and high efficiency. A novel QR color-code recognition method combining global and local features is proposed in [63]. In the method, the position of the QR code is detected through the global features, and the local features are used to identify the contents of QR codes. Although the system extend the detectable range to 15m, it cannot well detect obstacles below the waistline and is easily suffered from illumination change or motion blur.

*B. Navigation*

Sighted people can understand what is important for spatial learning and navigation through their vision, but the blind individuals may suffer from finding the path due to lack of vision [64]. Blind navigation works to provide the critical environmental information for visually impaired people. It is designed to allow the blind to travel effectively and safely in known or even unfamiliar environments by dynamically planning paths based on their current location and destination. Therefore, blind navigation should include two crucial factors: spatial location and spatial orientation, as shown in Fig. 7.

*1) Spatial Location :* Location is one of the significant factors in blind navigation, which can tell users where they are now [65]. As we all know, it is not easy to obtain the precise location with pure vision technology, so other assistive technologies are generally used to improve the accuracy. In [66], these positioning approaches are roughly divided into four categories: direct sensing, dead reckoning, triangulation, and pattern recognition, which will be elaborated in this subsection.

Direct sensing localization, such as radio frequency identification (RFID), Bluetooth beacons, and barcodes localization, can directly provide the user's location by sensing identifiers or marks pre-installed in the environment. Fernandes et al. used RFID as the main technology to develop a navigation system for visual impairments [67], which can guide blind people in unfamiliar environments by providing them with the geographical context. Ding et al. integrated a portable RFID reader into a mobile phone or a blind cane to build a blind navigation system [68], where the blind can understand their location and surroundings. However, this method needs to install many tags, which greatly increases the cost. In [69], Baus et al. proposed a hybrid navigation system with IR beacons that broadcast a unique ID to recognize the user's location. But it requires the user to be in the line-of-sight, and it strongly suffers from sunlight. A ubiquitous indoor navigation system with Bluetooth beacons is developed in [70] to help the blind find out the optimal path for destination. However, the system fails for fast walking users due to the communication delay. Also, a wearable system is developed to determine the people's location by reading the barcode on the information signs around the campus [71]. But it requires users to find each barcode and then scan it, which is unfriendly to the visually impaired. In addition, all the above technologies have to employ additional devices to sense the identifier, which further limits their application.

Dead reckoning localization generally refers to estimating one's current location using previously known locations [72]. Its updates involve tracking motion components (i.e. speed, acceleration, and heading) and driving duration. As the user moves, the dead reckoning localization uses the odometer readings to estimate the user's location. The odometer reading can be obtained by a combination of equipment such as acceleration sensors, magnetometers, compasses and gyroscopes [73]-[78], it can also be obtained by using a specific walking pattern such as the user's average walking speed [79]. Ribeiro et al. designed an auditory augmented reality system for visually impaired, which can estimate the location of users and obstacles [74]. Compared with the direct sensing localization, this approach costs less because it does not need to install a lot of identifiers. However, dead reckoning localization is susceptible to disturb from cumulative errors since position estimation is a recursive process.

Triangulation localization gets the user's position by triangulating the tags pre-installed in known places. Unlike direct sensing localization that locates users by sensing a unique identifier, the triangulation localization adopts multiple identifiers to locate users. *Lateration* uses the distance between the user and at least three known points, whereas *Angulation* uses the angular measurements from at least three known points for the user to calculate locations [80], as shown in Fig. 9, where $R$ represents the user, $B_{i,\ i=1,2,3}$ denotes beacons, $\alpha_{i,\ i=1,2,3}$ denotes the angles, and $\theta$ represents the user's orientation [81]. These angles are used by a triangulation algorithm to obtain the user's position. Global positioning system (GPS), the most representative outdoor positioning technology, adopts satellite-based trilateration to determine users' locations. It analyses the periodic signals sent by each satellite to calculate the user's location information such as latitude, longitude, and altitude [82], [83]. Velázquez et al. developed a wearable outdoor navigation system based on GPS for visually impaired [84]. In this system, real-time GPS data provided by a smartphone is processed to locate the user's location. The GPS data is almost available in any outdoor places on earth whether it is cloudy or rainy. However, when it comes to the situation inside buildings or between tall buildings, the performance of GPS will be greatly reduced.

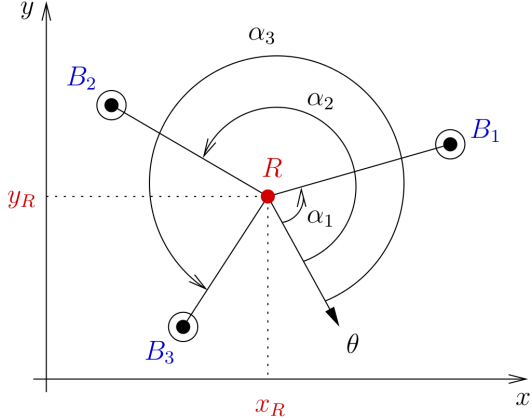

Fig. 9. Triangulation setup in the 2-D plane [81].

Pattern recognition localization refers to the position is obtained through matching the current image against the pre-collected environment map. In [66], this localization technology is grouped into two categories: signal distribution mapping localization and computer vision-based mapping localization. The signal from WLANs access points is an example of the signal distribution mapping localization [85], [86], and it only requires some base station to locate the user [87]. Computer vision localization can determine the user's position by matching the real-time images against the image database with known location [88]. Poggi and Mattoccia developed a wearable mobility aid that achieves the blind guidance using deep learning [25]. However, this aid can only provide an approximate distance information (such as how far the obstacle is) rather than precise location. Then, some researchers started combining visual sensors with the positioning technology, such as GPS [89], to promote the positioning accuracy.

Of course, there are some new positioning technologies that can serve the blind, which are not in the above four categories, such as sound localization [31]. But there are less related researches, they are not discussed in the paper.

*2) Spatial Orientation :* Orientation is another crucial factor in blind navigation, which relates to where the user would like to go. It refers to the process of tracking position and heading in the environment while traveling from one location to another [90]. According to the different generation methods, orientation can be broadly grouped into three categories: step-by-step, predefined-way, and way-finding.

Step-by-step orientation belongs to the kind of purpose-less navigation. It generally determines the next direction by analysing the current position and environment. Bai et al. developed a novel multi-sensor orientation system for visually impaired [32]. This system uses the region of interest to find the candidate traversable directions and further determines the optimal forward direction based on the multi-sensor obstacle avoidance algorithm. Filipe et al. introduced a 3D camera based indoor navigation system for blind people [91]. The system recognizes the forward direction by extracting six vertical lines at predefined locations in the depth image. In addition, Literature [92] used a weak-calibrated stereo camera to design a wearable navigation system. The system can assist the blind to avoid stationary and dynamic objects and find a safe walking region. However, the step-by-step orientation cannot draw a specific route for the blind, and it is mainly used to navigate in a familiar environment. Therefore, this type of orientation is insufficient obviously to meet the requirements of the blind.

Predefined-way orientation belongs to the purposeful navigation. The orientation guides the visually impaired people to their destination based on a predefined route. Blind lane is the most representative example of the predefined-way orientation. It helps visual impairments determine orientation through two types of bricks paved on the ground. Ding et al. presented an RFID-based virtual blind lane that can guide blind people to where they want to go by recognizing pre-installing cue tags buried near crossings, roads, and public sites [93]. Noman et al. introduced a novel assistive robot for visually impaired [94]. The robot can sense predefined lines and help blind people navigate indoors with infrared sensors. Predefined-way orientation can also serve some large public places. For example, when a user is in a supermarket and wants to go from one place to another, a predefined nearest route can be found from the database. However, such orientation can only provide some fixed routes, which limits its applied scenarios. Currently, it is mainly used in some special sites.

Way-finding orientation, which also belongs to purposeful navigation, is currently the most representative orientation approach. It is dedicated to considering the user's needs and customizing paths in real-time no matter where the users are and which direction they want to go. For outdoor environments, GPS has always been the most mature orientation approach [95]. Literature [96] did research on GPS-based mobile device applications for blind navigation in outdoor environments. This study introduces many outdoor navigation systems that can enable the blind to use GPS for way-finding in unknown environments. Since GPS signals cannot work well in indoors, it is important to find another technology that ensures accurate navigation in indoor environments [97]-[99]. Nguyen et al. used the visual odometry technique to develop a way-finding system based on a mobile robot [100]. This system can build reliable travel routes for the blind in indoors. Also, Cheraghi et al. developed a beacon-based indoor way-finding system for the blind and disoriented [101].

There are also some way-finding systems that can be operated in both indoor and outdoor environments. A GPS and RFID integrated navigation robot was designed in [102] for the visually impaired people. The robot can guide the blind to a predefined place, or build a new route on-the-fly for later use. In addition to the existing mature technologies, some researchers are working on developing new way-finding algorithms. Aria proposed a novel algorithm for finding the shortest path that can be implemented on a digital way-finding map [103], while Zhang et al. developed a hybrid genetic and ant colony algorithm to find the shortest path in the dynamic traffic networks [104].

Another point, often overlooked in the literature, needs special attention: the angle of the steering. It is well known that sighted people can correct minor orientation errors by visually inspecting the surroundings. However, for visually impaired people, a slight incorrect orientation may cause them to get lost, because they are not able to recover the orientation errors. In blind navigation, it is obvious that simple orientation instructions such as "turn left" or "45 degrees left" are not enough. Although Ahmetovic et al. studied the effect of steering errors in turn-by-turn navigation for blind people [105], few people have paid attention to this issue. Researches on steering angles and orientation errors in blind navigation still need more attention.

## C. Interaction

For a qualified VAS, it is crucial to interact with the users to receive instruction and provide warning or surrounding information, which directly affects the users' safety. Generally, a friendly interaction with visually impaired can be divided into two parts: input and feedback.

*1) Input :* A proper input ensures that the assistive system can accept the blind's instructions such as way-finding or object-searching. At the early stages, some handheld input devices are adopted to help the blind interact with the environment [66]. For example, Nakamura et al. developed a walking navigation system that used push-button switches to interact with blind [106]. A travel aid that used touch screens to receive instructions is designed in literature [107], while the keypads is adopted as the input to interact with the blind people in literature [108]. As we all know, it is important for the blind to keep their hands free. However, such approaches require the blind to hold these devices in their hands all the time, which is unfriendly to visually impaired people.

Recently, with the development of speech recognition, some STT/TST (speech to text/text to speech) products such as Sphinx [109] or Hidden Markov Model Toolkit [110] are widely used in the field of blind assistance [111]. Deshpande and Shriram presented a real-time text detection and recognition system for blind people [61]. The system can recognize the text in the environment and convert it to an audio output. Harsur and Chitra designed a voice-based navigation system for blind people [112]. The system adopts a Pocket Sphinx and Google API to transform the speech into the text, otherwise converts the text to speech with Espeak tools. Although these approaches are easily affected by the ambient sounds in the environment, they don't take up user's hands.

*2) Feedback :* Feedback refers to the process in which the assistive system returns a response to the user. There are currently three representative ways to provide feedback to visually impaired people: visual, audio and haptic feedback. Visual feedback adopts visual cues to respond to the user, which is a relatively rare type of feedback for the blind. It is mainly suitable for partially sighted people to get more detailed information [113]. Generally, the display is the most common tool for assistive system based on visual feedback. For example, Hicks et al. designed a head-mounted visual display to help the partially sighted individuals to navigate [114]. The system correlates the distance with the brightness on the display to enhance the user's visual ability. However, visual feedback can hardly work for the people who are completely blind.

Audio feedback is currently the most used feedback, which delivers the information to users through the sound. Kang et al. introduced a novel obstacle detection system that devotes to interact with the blind by converting the outputs into acoustic patterns [115]. In [116], Stoll et al. transformed the distance into sounds of different intensities to warn users to avoid obstacles. In addition, literature [117] aimed to helping the blind avoid obstacles by mapping the depth image to semantic speech, whereas a vision-based assistance system is developed in [118], which perceive the road situation by listening to the stereo musical sound. Furthermore, Bai et al. developed a sound synthesis module specifically for the blind, which can generate three kinds of guidance signals: recorded instructions, stereo sound tones, and different frequency beeps, as shown in Fig. 10 [32]. However, audio feedback is language-dependent. Sometimes, lengthy speech may prevent the blind people from hearing the key information.

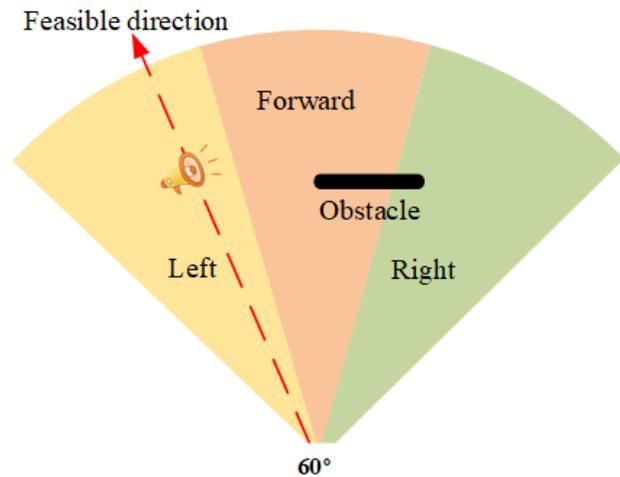

Fig. 10. Guiding sound synthesis module.

Therefore, feedback should not overwhelm the user's senses while providing necessary information [119]. Haptic feedback is also a good choice for blind to interact with the assistive system. There are two types of haptic feedback: active haptic feedback and passive haptic feedback. Braille is a good example of the active haptic feedback. Wang et al. introduced an independently wearable navigation system for visually impairments [120]. In this system, a handheld module with a 10-cell braille display was designed to offer high-level feedback in a discrete way. Pissaloux et al. developed a new framework for cognitive mobility of blind people [121]. A TactiPad consisting of a matrix of 8 × 8 taxels is used in the framework to get the acquisition of spatial knowledge. However, braille feedback requires longer response times and only works for people who have learned braille. Passive haptic feedback generally adopts some vibrators to response to the blind user. For example, Johnson and Higgins used a tactor belt containing 14 laterally spaced vibration motors to guide blind people to determine the right direction [122], while Mann et al. adopted a vibro-tactile helmet to present the way-finding result [123]. Haptic feedback has the advantage of much less interference with perceiving the environment. But it is difficult to deliver complex information and sometimes requires more training and focus. In addition, prolonged haptic feedback can also drain the battery [66]. There are also some researches that fused the multiple types of senses to provide better feedback performance. Hashimoto and Takagi combined haptic and audio cues to propose an audio-tactile graphic system based on smartphones to help the blind people access the surrounding information [124].

## V. Typical Application

In the last decade, many vision-based assistive systems were proposed to help the visual impairments travel freely. According to the different application scenarios, these systems can be divided into two cases: outdoor applications and indoor applications.

### A. Outdoor Applications

Currently, almost all the outdoor assistive systems are developed based on the GPS, since it can obtain a person's location in real time from almost anywhere on the ground of earth. The positioning accuracy of GPS is about 1m to 10m, which is enough for normal people. But for the visually impaired, such accuracy is not enough. Therefore, some researches fused vision technology and GPS to further improve positioning accuracy.

Meers and Ward combined vision and GPS to propose an obstacle avoidance and outdoor navigation system [125]. The user's location is roughly obtained by GPS in the system, while the 3D surroundings is perceived by measuring the disparity between the stereo images. In [126], a real-time navigation and monitoring system is proposed. The system introduces a remotely real-time navigation and assistance mode based on GPS, which can help the blind ask somebody for help in emergencies through a video taken by the RF (range finder) camera. Brilhault et al. fused GPS and vision to develop an assistive system that can locate the blind in urban environments where GPS signals are degraded [89]. In the system, the vision module can help the blind refine objects location along the route, which is rough built by GPS.

### B. Indoor Applications

GPS technology has promoted greatly the quality of the blind's life in outdoors, so far no technology that has convinced everyone can be worked for indoor blind navigation [27]. Current indoor systems rely on various technologies to guide the blind people. According to the environmental dependence, these technologies can be broadly divided into two types.

The first is to use portable assistive devices to help visually impaired avoid obstacles, locate and travel safely [127]. In [88], a stereo camera-based navigation system is designed to help visually impaired orient themselves in indoors. The system uses a client-server-like approach to help blind people perceive the 3D environments and find the optimal direction. A vision-based 3D perception system is proposed in [128] to help blind people navigate in working and living environments. The system creates a 3D representation of the surrounding space, which can help users understand the changes in 3D space and chose the proper direction.

The second type is to integrate sensors or signs into the environment to help the visually impaired to move freely in the indoors. Chumkamon et al. presented an RFID indoor navigation system for visually impaired [130]. The system relies on the location information on the tag to locate the user and find the shortest route. However, the study shows the system also has some latency issues. Ma and Zheng introduced the UWB wireless technology to the blind assistive systems, and proposed a high precision indoor navigation system that mainly includes mobile phone, location base stations, router, electronic tags, wireless transceiver, and computer, as shown in Fig. 11 [131]. Although the system effectively improves the positioning accuracy, it requires that the computer and the mobile phone must be on the same wireless network to ensure the normal communication between the mobile phone and computer. In [132], the QR codes scanning technique is used to find the optimal route for the blind. In this system, QR codes containing the position information are placed on the floor. Whenever the QR code is scanned, the system will find the location information implied in the QR code, then the optimal route can be planned for the blind. However, users have to find the QR codes by themselves, which is unfriendly for the visually impaired people.

As a matter of fact, all the above systems require large investment for installation and maintenance [133]. In order to reduce costs, Ahmetovic et al. adopted the pedestrian dead reckoning (PDR) to improve the previous system [127]. The new system achieves a better localization accuracy using only a few beacons. Also, Dockstader et al. developed a multiple camera-based system for positioning and tracking multiple persons in motion [134], as shown in Fig. 12. Although the system is not directly designed for the blind, it can certainly be used to serve the blind people.

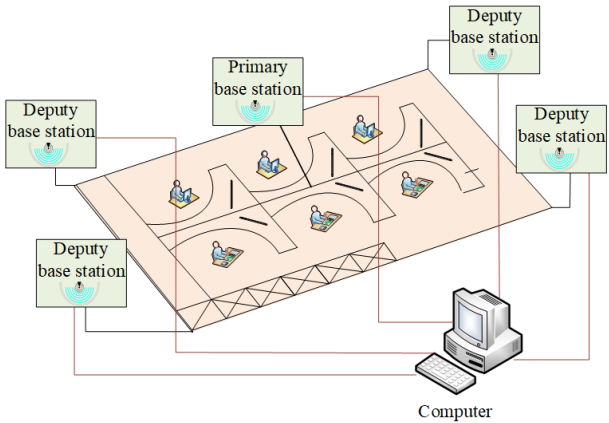

Fig. 11. The high precision blind navigation system proposed in literature [131].

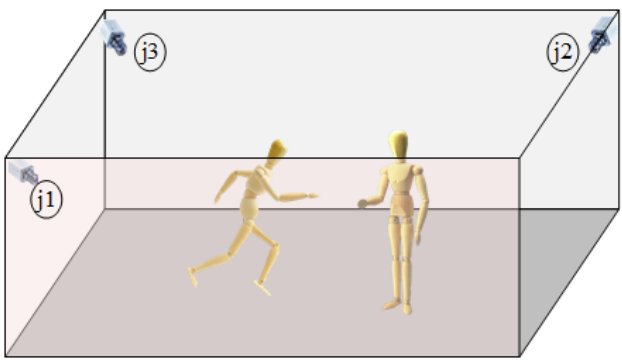

Fig. 12. Fig. 12 Three-dimensional view of the scene in literature [134].

## VI. Discussion

In the last decade, significant progress has been achieved on the assistive systems for visually impaired people, TABLE II provides an overview of some of the latest assistive systems discussed above. There are still some challenges that need to be addressed, which reveal the future direction of the blind guidance to some extent.

- Assistive systems for the blind should be able to detect, recognize and learn new objects in unexplored environments. Current algorithms can only classify and recognize objects that are pre-stored in the database, but cannot identify unrecorded objects. An ideal assistive system should be able to help the blind identify new things. So it is still a serious challenge to build a assistive system with self-learning function.
- A portable 3D vision sensor with high resolution is urgently required. 3D cameras can capture some information that 2D cameras cannot obtain, such as depth information, but low-precision or low-resolution 3D cameras often produce some blurry images, especially in motion. Apart from hardware innovations, algorithms such as super-resolution reconstruction can also be used to improve the image's quality. Existing literature shows that such algorithms can effectively improve the quality of pictures, but the time cost should be taken into account.
- A portable high-precision positioning and navigation technology, which is suitable for indoors and outdoors, remains to be discovered. While GPS can be used for outdoor positioning and way-finding, most indoor way-finding methods still need additional the equipment such as RFID or BLE. It is necessary to develop a new portable navigation technology for both indoors and outdoors.
- There are also some issues that need to be addressed with regard to interaction. The assistive system must not cause any interference in daily activities, otherwise it will be regarded as an obstacle. However, any feedbacks will have different degrees of adverse effects. Visual feedback is only available for the partially sighted people, audio feedback may prevent the blind from hearing environmental information, active haptic feedback requires longer response times, passive haptic feedback could cause user misunderstanding with consequent vibration. Therefore, it may be necessary to establish a multi-mode selection mechanism so that users can choose the proper model.
- The assistive system may cause a certain degree of environmental damage. The future system should minimize the interference to the environment while meeting the needs of the blind. In addition, most devices have only been verified by sighted people at specific venues. It is not certain that these devices can work well in the blind people. So the new system should be promoted to the actual scene to check the rationality of the equipment.

It is believed that the future of blind assistance systems should follow the direction of developing integrated products that will take advantage of different technologies for efficiency and robustness. Multi-technology fusion does not mean complexity, and it should follow the principles of low cost and portability.

## VII. Conclusion

Nowadays, the need of assistive systems for the visually impaired is greater than ever before as human life expectancy increases. This article presents a comprehensive overview of the state-of-the-art assistive systems, especially VAS, for visually impaired, including five parts: 1) haraware sensor, 2) object recognition and avoidance, 3) user positioning and orientation, 4) interaction, 5) typical applications. Also, different technologies or methods are compared in this paper, and their strengths and weaknesses are analysed in terms of cost, accuracy, and performance. Then, this paper discusses the current challenges faced by assistive technologies for the blind.

TABLE II
Overview of some of the latest assistive systems

"✓": the tool/technology was used, "×": the tool/technology was not used, "-": the tool/technology was not mentioned.

| Author | Year | Reference | Input | | | | | | | | |
|---|---|---|---|---|---|---|---|---|---|---|---|
| | | | Traditional tools | | Electronic sensors | | | Visual sensors | | | |
| | | | Cane | Dog | Ultrasonic | Laser | IR | Monocular | Omnidirectional | RGB-D | Stereo |
| Al-Fahoum | 2013 | [19] | ✓ | × | × | × | ✓ | × | × | × | × |
| Hu | 2014 | [38] | × | × | × | × | × | × | ✓ | × | × |
| Xiao | 2015 | [31] | × | × | × | × | × | × | × | ✓ | × |
| Dang | 2016 | [18] | × | × | × | ✓ | × | ✓ | × | × | × |
| Poggi | 2016 | [25] | ✓ | × | × | × | × | × | × | ✓ | × |
| Bai | 2017 | [32] | × | × | ✓ | × | × | × | × | × | ✓ |
| Noman | 2017 | [94] | × | × | ✓ | × | ✓ | × | × | × | × |
| Chuang | 2018 | [26] | × | ✓ | × | × | × | ✓ | × | × | × |
| Mante | 2018 | [34] | × | × | × | × | × | ✓ | × | × | × |
| Lee | 2018 | [63] | × | × | ✓ | × | × | ✓ | × | × | × |
| Rahman | 2019 | [33] | × | × | × | × | × | ✓ | × | × | × |
| Hakim | 2019 | [48] | × | × | ✓ | × | × | × | × | ✓ | × |

| Reference | Obstacle detection | | Spatial location | | | |
|---|---|---|---|---|---|---|
| | Avoidance | Recognition | Sensor | Dead reckoning | Triangulation | Pattern recognition |
| [19] | ✓ | × | ✓ | × | × | × |
| [38] | × | ✓ | × | × | × | ✓ |
| [31] | ✓ | × | × | × | ✓ | ✓ |
| [18] | × | ✓ | ✓ | × | × | × |
| [25] | × | ✓ | × | × | ✓ | × |
| [32] | × | ✓ | ✓ | × | × | × |
| [94] | ✓ | × | ✓ | × | × | × |
| [26] | - | - | × | × | × | ✓ |
| [34] | - | - | × | × | × | ✓ |
| [63] | ✓ | × | ✓ | × | × | ✓ |
| [33] | ✓ | × | × | × | × | ✓ |
| [48] | ✓ | × | - | - | - | - |

| Reference | Spatial orientation | | | Interaction | | | Application | | |
|---|---|---|---|---|---|---|---|---|---|
| | Step-by-step | Pre-defined | Way-finding | Visual | Audio | Haptic | Outdoor | Indoor | |
| | | | | | | | | Portable | Integrated |
| [19] | ✓ | × | × | × | ✓ | ✓ | ✓ | ✓ | × |
| [38] | × | ✓ | × | - | - | - | × | ✓ | × |
| [31] | × | × | ✓ | × | ✓ | ✓ | ✓ | ✓ | × |
| [18] | ✓ | × | × | × | ✓ | × | × | ✓ | × |
| [25] | × | × | ✓ | × | ✓ | ✓ | ✓ | × | × |
| [32] | ✓ | × | × | ✓ | ✓ | × | × | ✓ | × |
| [94] | × | ✓ | × | × | ✓ | ✓ | × | × | ✓ |
| [26] | × | ✓ | × | - | - | - | ✓ | × | ✓ |
| [34] | - | - | - | × | ✓ | ✓ | × | ✓ | × |
| [63] | ✓ | × | × | × | ✓ | ✓ | × | ✓ | × |
| [33] | ✓ | × | × | - | - | - | × | ✓ | × |
| [48] | ✓ | × | × | × | ✓ | × | × | ✓ | × |

Among different kinds of assistive technologies, VAS will continue to attract attention and help visually impaired in an unprecedented way. Moreover, multiple technologies can be integrated to provide efficient and robust aids to visually impaired people, which have a great potential in the future market.